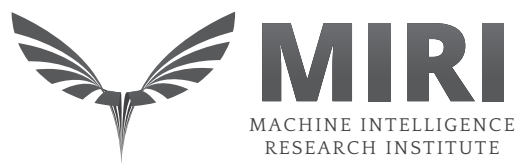

# A Comparison of Decision Algorithms on Newcomblike Problems

Alex Altair
*MIRI Research Fellow*

**Abstract**

When formulated using Bayesian networks, two standard decision algorithms (Evidential Decision Theory and Causal Decision Theory) can be shown to fail systematically when faced with aspects of the prisoner's dilemma and so-called "Newcomblike" problems. We describe a new form of decision algorithm, called Timeless Decision Theory, which consistently wins on these problems.

# 1. Three Decision Algorithms

Decision theory typically analyzes two main decision algorithms known as Evidential Decision Theory (EDT) and Causal Decision Theory (CDT).[1] When formalized using Bayesian networks, CDT and EDT can both be shown to return the wrong results on certain types of decision problems known as Newcomblike problems (named after the original "Newcomb's problem"). Systematically achieving the correct results when faced with these problems is essential for decision-theoretic qualities like dynamic consistency and the ability to model the self.[2] Much work is dedicated to finding algorithms that maximize expected utility on these problems (for example, Wedgwood [2011]). In this paper we will consider such an alternative algorithm proposed by Yudkowsky (2010), called Timeless Decision Theory (TDT), which was inspired by Douglas Hofstadter's notion of "superrationality" (Hofstadter 1985).

All decision algorithms described in this paper can be formalized to use Bayesian networks (after Pearl [2009]) introduced in section 2. Using these, we demonstrate the undesirable results returned by EDT on a version of the so-called medical Newcomb's problem (section 3) and by CDT on the prisoner's dilemma (section 4). In section 5 we introduce the innovation in TDT, and section 6 uses it to justify cooperating in the one-shot prisoner's dilemma against similar players. Section 7 reviews some open problems that TDT does not solve.

Decision theories are instantiated in *agents*; that is, in physical systems that observe the world and implement a decision algorithm. EDT, CDT and TDT will therefore be part of a system that has other capabilities, such as inferring causal networks from observations, applying a utility function to these networks, and identifying parts of the network that represent the agent itself and possible actions. Parts like these are necessary to implement any decision algorithm but will not be formally described in this paper. As part of an agent, decision algorithms are passed a formal description of a decision problem, here as a causal network, and output one of the possible actions as the decision. All three attempt to select the action $\alpha \in A$ that maximizes expected utility. That is, they return:

$$\alpha = \arg\max_{a \in A} \sum_{o \in O} U(o|a) P(o|a) \qquad (1.1)$$

---

1. EDT was introduced by Jeffrey (1983). See Joyce (1999) for a comprehensive reference on CDT.

2. A "correct" result is defined as one that maximizes expected utility, which itself is judged in part by our intuition. Formalizing the rightness criterion to match with our educated intuition is precisely the goal of decision theory.

where $P(o|a) = P(oa)/P(a)$ as usual.[3] The set $A$ contains all actions the agent can take. They are mutually exclusive and exhaustive; there are no possible actions except those in $A$. The set $O$ contains all outcomes, also mutually exclusive and exhaustive. Given that the agent takes action $a$, $P(o|a)$ is the probability of outcome $o$ and $U(o|a)$ is the utility of outcome $o$ (notation by analogy with conditional probabilities). The decision algorithms disagree on how to compute the conditional probability $P(o|a)$ of an outcome given an action. Note that, when analyzing these conditionals, the algorithms have not yet chosen an action. Furthermore, since they can only choose one action, most actions will not be chosen. Yet the agents must analyze the conditional probability $P(o|a)$ for all actions. It is partly this intuitively confusing fact that leads to different proposed decision theories. Those discussed here differ in how they modify the causal network from which they use the resulting probability distribution.

## 2. Causal Bayesian Networks

CDT and EDT are usually distinguished by means of their interpretations of probability in decision problems. EDT is given the standard notation and interpretation of $P(o|a) = P(oa)/P(a)$. In discussion this is often translated as the probability of $o$ given that you receive the news that $a$ occurred. It is just how a Bayesian reasoner would calculate the conditional if it were about someone else's decision. CDT has many formalizations, most of which define a kind of two-variable probability function $P(a, o)$ to distinguish it from EDT by recognizing the causality communicated between $a$ and $o$.

With Bayesian networks we can be more precise. We will formalize all three decision theories as algorithms that use the same expected utility maximizing calculation, but which differ in how they modify the causal network. Here we present a bare-bones introduction to Bayesian networks. For a full introductory exposition see Pearl (2009); a briefer non-introductory piece is Koller and Friedman (2009, chap. 21).

Let $X_1, X_2, \ldots, X_n$ be a collection of random variables, where $x_i$ is a specific value of $X_i$. A graphical model is a graph in which each of the $X_i$ is represented by a node. If it is also a directed acyclic graph with a joint probability distribution satisfying equation 2.1, then it is a Bayesian network. $PA_j$ is a vector-valued variable of the set of parents of $X_j$, defined as those nodes in the graph that have directed arrows to $X_j$. If these arrows represent causal dependence, then it is a *causal Bayesian network*. All graphs in this paper will be causal Bayesian networks; hereafter we may simply refer to them as networks. Given a network, we can write a probability distribution

$$P(x_1 x_2 \ldots x_n) = \prod_i P(x_i | pa_i) \tag{2.1}$$

where $pa_i$ is a generic value of $PA_i$. All other probabilities can be derived from this. Additionally, if we want to model a counterfactual value of $x_i$, we can do so by removing the arrows from $PA_i$ to $X_i$ and setting $P(x_i | pa_i) = P(x_i)$.

---

3. In general we will be relaxed about probability notation when it is clear from context. $P(x)$ may refer to a distribution over many possible values of $x$, or it may refer to the probability of a specific proposition $x$.

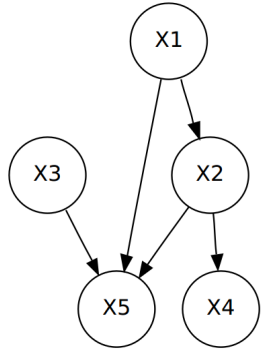

Figure 1: An example Bayesian network.

Let us build some intuition by reviewing this with an example network, shown in figure 1. A naive distribution for this graph from the chain rule would be

$$P(x_5 x_4 x_3 x_2 x_1) = P(x_5 | x_4 x_3 x_2 x_1) P(x_4 | x_3 x_2 x_1) P(x_3 | x_2 x_1) P(x_2 | x_1) P(x_1). \tag{2.2}$$

But by using equation 2.1 above, we can simplify this to

$$P(x_5 x_4 x_3 x_2 x_1) = P(x_5 | x_3 x_2 x_1) P(x_4 | x_2) P(x_3) P(x_2 | x_1) P(x_1). \tag{2.3}$$

These independencies have fundamental implications for reasoning with Bayesian networks. For instance, the value of $x_5$ is *caused* by $x_1$, $x_2$ and $x_3$, but knowing $x_4$ gives you information about $x_2$, which gives you information about $x_5$. Similarly, if you know $x_2$, then knowing $x_4$ does not give you any *more* information about $x_5$. The following equations also hold for figure 1, and can be derived from equation 2.3 using basic laws of probability.

$$P(x_1 | x_2 x_4) = P(x_1 | x_2) \tag{2.4a}$$
$$P(x_4 | x_5 x_2) = P(x_4 | x_2) \tag{2.4b}$$
$$P(x_4 | x_2 x_1) = P(x_4 | x_2) \tag{2.4c}$$

It can be said that $x_5$ "screens off" $x_3$ from $x_2$, and so on. Conversely, the following expressions can't be simplified in a similar way: $P(x_1 | x_4 x_5)$, $P(x_3 | x_1 x_4)$, $P(x_2 | x_3 x_5)$.

In our networks, one node represents the agent's possible actions. We will call this the *decision node*, and it will be denoted by a double boundary. The possible values of this node will constitute the set $A$ in the expected utility equation 1.1. Furthermore, the set $O$ from 1.1 will consist of the set of possible states of the whole network; in other words, the domain of the joint probability distribution.

When computing $P(o|a)$ in the expected utility, EDT uses $P(o|a)$ from the causal network given to it. CDT calculates a decision as an intervention in the network. That is, it calculates each $P(o|a)$ as if the decision node is determined. This severs the causal connections from ancestors of the decision node, and *then* computes $P(o|a)$ from the network. TDT *adds* a logical node above the decision node that represents the computation that is TDT's decision process. That is, a TDT agent will model its decision not just as a physical process with some output, but as an algorithm, whose output is not known and is therefore modeled by subjective probability. If any other nodes represent this same algorithm, they will have the same output; therefore TDT also adds a directed edge from the logical node to those other nodes. By adding the logical node as a parent to these, TDT recognizes their connection in reality. The logical node is then considered to be the decision node, not just the node representing a specific instantiation of the TDT agent. TDT then analyzes this network like CDT would: it severs connections above the decision node, then calculates $P(o|a)$ from the network.

## 3. Toxoplasmosis Problem

The toxoplasmosis problem is a scenario that demonstrates a failing of EDT and a success of CDT.[4] *Toxoplasma gondii* is a single-celled parasite carried by a significant fraction of humanity.[5] It affects mammals in general and is primarily hosted by cats. Infection can have a wide range of negative effects (though most show no symptoms). It has also been observed that infected rats will be less afraid of cats, and even attracted to cat urine. Correlations have been shown between psychiatric disorders and toxoplasmosis, and it has been speculated (but not tested) that the disease may cause people to be more risk taking, and attracted to cats. Neurological mechanisms have been proposed (Flegr 2007).

---

4. Historical counterexamples to EDT include Solomon's problem (Gibbard and Harper 1978), and the smoking lesion problem (Egan 2007). We use toxoplasmosis as a real-life variant that does not use concretes that strongly conflict with common knowledge and intuition. In the smoking lesion problem, it is a premise that smoking does not cause cancer. Humans now know the fact that smoking causes cancer (Kuper, Boffetta, and Adami 2002), and this can make it difficult to query one's intuitions about decision principles only.

5. We labor here to claim no medical facts or advice. No guarantee is given of the truth of clinical statements; our only purpose is to inform the reader of the realism of decision theoretic problems.

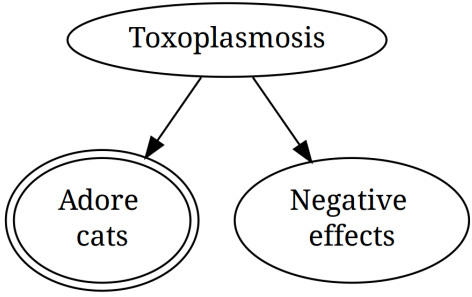
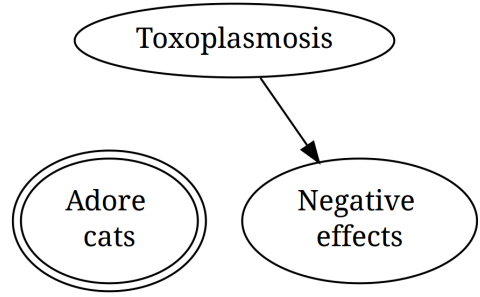

Figure 2: Bayesian network of the toxoplasmosis problem.

Figure 3: CDT's intervention modification of the network.

Whether or not these ideas are correct, they are now properly represented by nonzero subjective probabilities. For the sake of example, assume our decision agent believes in the validity of figure 2. Let $C$ stand for adoring cats and $N$ stand for having the negative effects of toxoplasmosis. Alternatives are denoted by affixing negation ($\neg$). Having toxoplasmosis increases the chances of the negative side effects, as well as increasing the chance of adoring cats. That is, $P(N|C) > P(N|\neg C)$. Note that $P(\neg N|C) = 1 - P(N|C)$ and therefore $P(\neg N|C) < P(\neg N|\neg C)$. Furthermore, let our agent have the following utilities, where $-B$ stands for the big negative utility from having toxoplasmosis symptoms, and $s$ for the small utility from adoring cats, with $B \gg s$.

$$U(N|C) = -B + s \qquad U(\neg N|C) = 0 + s \qquad (3.1)$$
$$U(N|\neg C) = -B + 0 \qquad U(\neg N|\neg C) = 0 + 0$$

EDT will assess the expected utility of adoring cats by taking conditional probabilities from the diagram, taking utilities above, and putting them in equation 1.1.

$$E(C) = P(N|C)U(N|C) + P(\neg N|C)U(\neg N|C) \qquad (3.2)$$
$$= P(N|C)(-B + s) + P(\neg N|C)(0 + s)$$
$$= \big(P(N|C) + P(\neg N|C)\big)s - P(N|C)B$$
$$= \big(P(N|C) + 1 - P(N|C)\big)s - P(N|C)B$$
$$= s - P(N|C)B \qquad (3.3)$$

$$E(\neg C) = P(N|\neg C)U(N|\neg C) + P(\neg N|\neg C)U(\neg N|\neg C) \qquad (3.4)$$
$$= P(N|\neg C)(-B + 0) + P(\neg N|\neg C)(0 + 0)$$
$$= -P(N|\neg C)B \qquad (3.5)$$

The agent will choose the action whose expected utility is higher; therefore an EDT agent will choose to adore cats if and only if

$$\frac{s}{B} > P(N|C) - P(N|\neg C). \tag{3.6}$$

The important part is not what the condition is, but that there is a condition at all.

The CDT agent severs causal connections from the decision node to its parents, and then computes the expected utility from the resulting diagram, in figure 3. CDT recognizes that, while knowing that someone adores cats is *evidence* that they have toxoplasmosis, the fact is already determined and therefore irrelevant. CDT's calculations look like this:

$$E(C) = P(N|C)U(N|C) + P(\neg N|C)U(\neg N|C) \tag{3.7}$$
$$= P(N)(-B + s) + P(\neg N)(0 + s)$$

$$E(\neg C) = P(N|\neg C)U(N|\neg C) + P(\neg N|\neg C)U(\neg N|\neg C) \tag{3.8}$$
$$= P(N)(-B + 0) + P(\neg N)(0 + 0)$$

Since the probabilities in both cases are the same, and the utilities for adoring cats are higher, adoring cats dominates and is better in all cases. This we know intuitively to be the correct answer; our decision does not affect whether we have toxoplasmosis, and so we should adore cats.

## 4. Prisoner's Dilemma

CDT also has areas of failure. The prisoner's dilemma is a classic decision problem in game theory. Two players are presented with a payoff matrix shown below.

Table 1: Prisoner's dilemma payoff matrix.

|   | C | D |
|---|---|---|
| C | $u_3$ ($u_3$) | $u_1$ ($u_4$) |
| D | $u_4$ ($u_1$) | $u_2$ ($u_2$) |

Let our decision agent be the player represented by the rows, and our opponent represented by the columns. The situation is symmetrical; what we describe for the agent is true for the opponent. The agent can choose to *cooperate* (C) or *defect* (D). The payoff matrix shows the utility received by the agent given the choices of both the agent and the opponent (with the opponent's received utility in parentheses). It is a prisoner's dilemma if the following ordering holds. (As a mnemonic, consider the indices as example utilities.)

$$u_4 > u_3 > u_2 > u_1 \tag{4.1}$$

The players are causally isolated, and each wants only to maximize their expected utility. Figure 4 represents the causal Bayesian network describing this problem.[6]

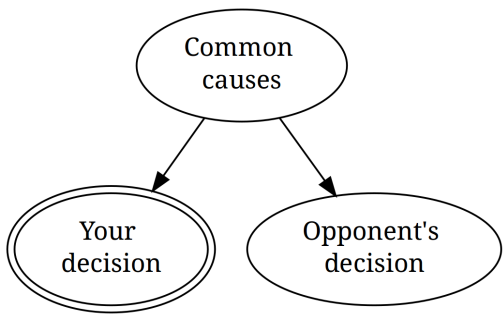 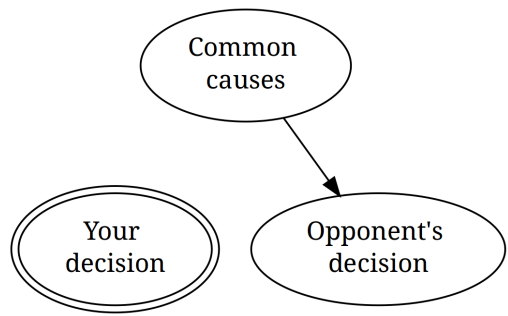

Figure 4: Causal Bayesian network of the prisoner's dilemma.

Figure 5: CDT's modification of the network.

The node "Your decision" can take on the variables $C_y$ or $D_y$ (you cooperate or you defect), and the node "Opponent's decision" can take on the variables $C_o$ or $D_o$ (opponent cooperates or opponent defects). In general, "Common causes" could consist of many possible causes with varying probabilities, from both players learning about decision theory in college, to being briefed before the game. The probability distribution could also be equal to the one where there is no common cause. CDT analyzes it as it did the toxoplasmosis problem: by severing connections to the decision node, and then solving equation 1.1 from the resulting diagram, shown in figure 5.

$$E(C_y) = P(C_o|C_y)U(C_o|C_y) + P(D_o|C_y)U(D_o|C_y) \tag{4.2a}$$
$$= P(C_o)u_3 + P(D_o)u_1 \tag{4.2b}$$
$$E(D_y) = P(C_o|D_y)U(C_o|D_y) + P(D_o|D_y)U(D_o|D_y) \tag{4.2c}$$
$$= P(C_o)u_4 + P(D_o)u_2 \tag{4.2d}$$

According to this analysis, $E(D_y)$ dominates $E(C_y)$; it is greater independent of the prior probabilities of your opponent's choices. Therefore CDT defects, always. If the CDT agent is facing an opponent that also runs CDT, this will lead to both receiving a utility of $u_2$, despite the fact that if they had both cooperated, they would both receive a utility of $u_3$.

In the standard prisoner's dilemma, it is difficult for human reasoners to come to a consensus over the right action. This is largely because you are told almost nothing about the opponent. A formal decision procedure such as EDT would not be able to make a decision without explicit probabilities over possible common causes.

---

6. It may be intuitive and reasonable to add another node, "Payoff," of which both your decision and the opponent's decision are parents. Since the states of the parents non-probabilistically determine the state of payoff, such a node is not necessary and clutters calculation.

## 5. Logical Nodes

To solve these Newcomblike problems in principle, TDT introduces two new stages of analysis.

1. All physical instantiations of the same logical fact (including results of computations) are recognized as having the same value by connecting them with a single parent node that represents the *abstract* logical fact.

2. TDT recognizes the output of its own decision algorithm as one of these logical facts.

To understand what these mean, and why a decision agent would need to model its own decision algorithm as a node, let us model a scenario involving two physical implementations of a calculation. The following is a modified version of a decision problem from Yudkowsky (2010).

Imagine that a group of pre-Columbian Mayan engineers requires more precision to build stable pyramid temples. After consulting with Incan quipu scholars, they devise a system of cords and pegs on axes to multiply up to twelve digits. To test their creation, they feed it ropes that encode the numbers 758439 and 990452: two factors to be multiplied.

Simultaneously, in Song Dynasty China, polymath cartographers work to automate the abacus. They too enter the factors 758439 and 990452 onto the input pegs of their creation.

Let us assume that both calculators correctly implement multiplication. Still, neither they nor we (unaided) know what the output of either machine will be, and therefore we model it as subjective uncertainty, which obeys the laws of probability. Figure 6 shows a naive causal Bayesian network formalization of this scenario. That is, there is no causal relationship between the two events. Presumably no cultural or information exchange has occurred between ancestors of these civilizations for thousands of years. We could have as easily framed it between alien species.

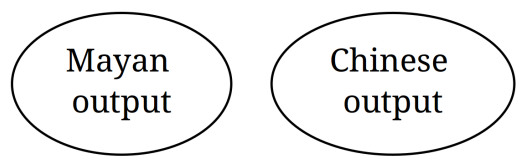

Figure 6: The naive causal network of the ancient calculators.

The joint probability distribution is equally simple:

$$P(c_M c_C) = P(c_M)P(c_C). \tag{5.1}$$

For simplicity, we will only consider the parity of the third digit of the product. Let *Maya-even* mean that the third digit of the Mayan calculator output is even, *Maya-odd* that it is odd, *China-even* mean that the third digit on the Chinese calculator is even, and *China-odd* that it is odd. Intuitively, we know that our subjective probability over these variables is uniform; that is, our guess on each variable being either even or odd is one-half. What is the probability that the third digit on the Mayan calculator is even, given that the third digit on the Chinese calculator is odd? The answer, of course, is effectively zero. But using the network in figure 6 we calculate the update a Bayesian reasoner would perform, upon learning the value returned by the Chinese calculator, as follows.

$$P(\text{Maya-even}|\text{China-odd}) = \frac{P(\text{Maya-even China-odd})}{P(\text{China-odd})}$$
$$= P(\text{Maya-even}) = .5$$

That is, our knowledge about one is independent from our uncertainty about the other. They are just as likely to display the same answer as a different one. This is because they have no causal effect on each other. Seeing an odd third digit on the Mayan calculator is no evidence of seeing an odd digit on the Chinese one.

Of course, this is the wrong answer.

Perhaps we've made a mistake in setting up our causal network, even by standard means. Perhaps the dependence is from the fact that we are specifying that they are both physical calculators, loaded with the same input. A physical correlation will come from the physical state of the calculators, so let's separate the state from the output of the calculators. The speculated causal network is shown in figure 7. The "calculator" nodes can take on an array of possible values, one of which is that the calculator's state is one that implements multiplication. The "output" nodes are the same as before, representing the parity of the third digit of the output of the calculators. The "Correlation" node represents dependency information from any possible physical correlation that we might be missing.

Given this network, we can test its validity by checking what predictions it makes. Let's say that we observe the state of both calculators, and determine that they both implement multiplication. What is now the probability that the Mayan calculator displays an even third digit, given that the Chinese calculator displays an odd one? Let $o_1$ be the output of the Mayan calculator, $s_1$ its physical state, $o_2$ the output of the Chinese

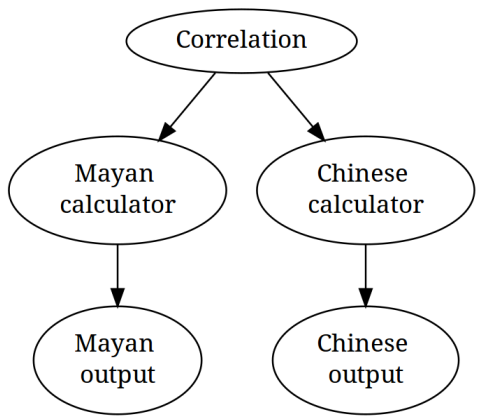

Figure 7: An attempt to account for the dependence between calculator outputs.

calculator, $s_2$ its physical state, and $c$ the value of the physical correlation:

$$P(o_1o_2|s_1s_2)P(s_1s_2) = P(o_1o_2s_1s_2)$$
$$= \sum_c P(o_1o_2s_1s_2c)$$
$$= \sum_c P(o_1|s_1)P(o_2|s_2)P(s_1|c)P(s_2|c)P(c)$$
$$= P(o_1|s_1)P(o_2|s_2)\sum_c P(s_1|c)P(s_2|c)P(c)$$
$$= P(o_1|s_1)P(o_2|s_2)\sum_c P(s_1s_2c)$$
$$= P(o_1|s_1)P(o_2|s_2)P(s_1s_2)$$
$$P(o_1o_2|s_1s_2) = P(o_1|s_1)P(o_2|s_2). \tag{5.2}$$

That is, once you know the states of the calculators, any dependence from physical correlation is screened off, and the outputs once again become independent. We still *do* have uncertainty about the outputs, because we still don't know $758439 \times 990452$. And the outputs *are* two different variables in the real world. We can't just combine the output nodes into one node; one of the calculators might implement multiplication, while the other one may have a malfunctioning carry mechanism. So what is wrong here?

It is intuitively clear that if we know that both calculators work correctly, and they are set to calculate the same product, the result is the same. Our uncertainty of the logical result—the unique and necessary product—is not the same as the kind of uncertainty in physical causality. Is there a way to represent it using the same formalism of causal networks?

Yudkowsky proposes that we add a node that represents the output of the calculation as a mathematical abstraction. To generalize, a *logical node* is a node in a Bayesian network that represents the network builder's subjective uncertainty about the result of a particular logical fact. This is added in figure 8. Now the calculator's outputs are made interdependent by this logical node representing the correct output. If we know one calculator's output, we can infer that the other's is the same using the usual Bayesian network updating algorithms.

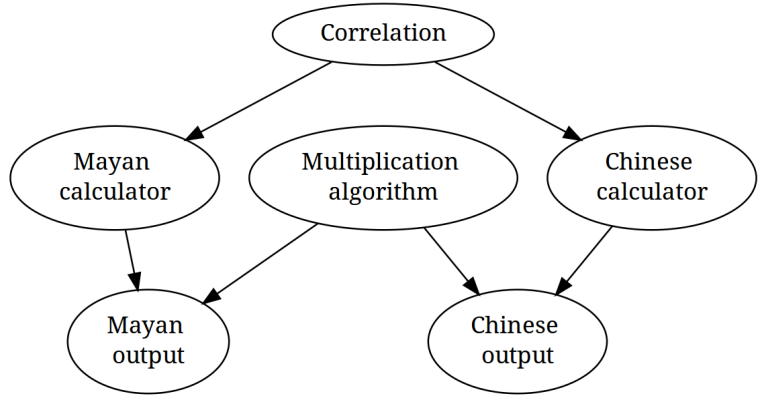

Figure 8: TDT's modification of the calculator problem.

## 6. Self-Modeling

The above example demonstrated how omitted logical nodes can create false beliefs, and how restoring the nodes can fix them. Omitted logical nodes also create problems in decision networks, as demonstrated by the above Newcomblike problems. Below we show the second step introduced by TDT: recognizing oneself as a decision algorithm.

Consider again the general prisoner's dilemma above. TDT's analysis begins with a modification of the network. It recognizes that both it and its opponent use decision algorithms, of whose output it is unsure. It places a node representing TDT's output above its own node, since it knows it uses TDT. It also connects the TDT node to its opponent's decision node, since its opponent could be using TDT. Above its opponent's node it also places a separate node for each decision algorithm its opponent could be using. Each of these decision-algorithm nodes can have a value of "Cooperate" or "Defect." Exactly how this is implemented depends on what decision algorithms the agent is aware of, and what probabilities it assigns to them being used by the opponent. Potential networks could look like figure 9. How the opponent is predicted to act is the result of how these possible decision algorithms act, and how likely they are to be used by the opponent. The predictions of figure 9 can be consolidated into a smaller diagram, shown in figure 10.

Like the others, the node "Not TDT" can have the value "Cooperate" ($C_n$) or "Defect" ($D_n$). Since we know that "Your decision" is deterministically chosen by "TDT," we can consider these nodes as one in calculations.[7] With this network, the expected utility calculation runs as follows.

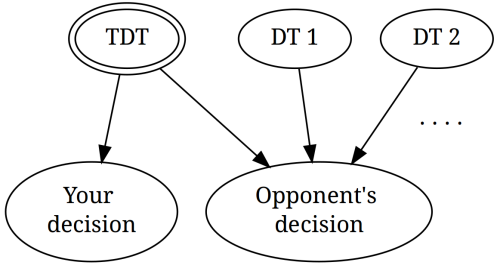

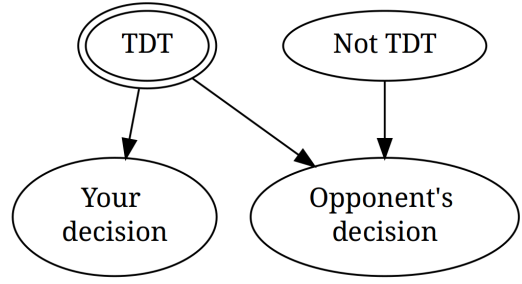

Figure 9: TDT's version of the prisoner's dilemma

Figure 10: Abbreviated diagram for prisoner's dilemma.

$$
\begin{aligned}
E(C_y) &= P(C_o|C_y)U(C_o|C_y) + P(D_o|C_y)U(D_o|C_y) \\
&= \big(P(C_o|C_yC_n)P(C_n) + P(C_o|C_yD_n)P(D_n)\big)u_3 \\
&\quad + \big(P(D_o|C_yC_n)P(C_n) + P(D_o|C_yD_n)P(D_n)\big)u_1
\end{aligned}
\quad (6.1a)
$$

$$
\begin{aligned}
E(D_y) &= P(C_o|D_y)U(C_o|D_y) + P(D_o|D_y)U(D_o|D_y) \\
&= \big(P(C_o|D_yC_n)P(C_n) + P(C_o|D_yD_n)P(D_n)\big)u_4 \\
&\quad + \big(P(D_o|D_yC_n)P(C_n) + P(D_o|D_yD_n)P(D_n)\big)u_2
\end{aligned}
\quad (6.1b)
$$

Unfortunately, no further simplification is possible. The dominance is gone, and whether cooperation or defection is better depends on the conditionals.

Though more information is needed to make a numerical judgment, in many cases it is reasonable to think that, for instance, $P(C_o|C_y) > .5$. That is, it seems reasonable to believe that your opponent will make choices similar to yours: that your opponent will reason similarly to you. If it is true that

$$P(C_o|C_y) \lessapprox 1, P(D_o|D_y) \lessapprox 1 \quad (6.2)$$

and

$$P(C_o|D_y) \gtrapprox 0, P(D_o|C_y) \gtrapprox 0 \quad (6.3)$$

then it is the case that $E(C_y) > E(D_y)$. If TDT believes the above conditions, it will cooperate. For instance if TDT knows that its opponent runs TDT, it will know that $P(C_o|C_y) = 1$ and will therefore cooperate. On the other hand, if we have no reason to think that our opponent's reasoning processes are similar to ours, then $P(o|y) \approx P(o)$ and $E(D_y) > E(C_y)$. That is, dominance is restored. For instance, if TDT knows its partner cooperates always, then TDT will defect.

---

7. Probabilities of 1 or 0 are not typical in Bayesian networks, but they are required to properly model some circumstances. Then one must be careful not to evaluate a probability conditioned on an impossible event. For more detail see Pearl (1988, chap. 3).

Adding these logical nodes can be considered an addendum to CDT. Others have published similar modifications of CDT. Spohn (2012, 2003) models Newcomblike problems with Pearl's networks and proposes adding nodes that represent intention. He recognizes these intention decisions as deterministic procedures. Spohn's intention nodes don't quite capture the same concept as logical nodes; Spohn (2003) fails to achieve mutual cooperation in the one-shot prisoner's dilemma, at which TDT succeeds. Drescher (2006) discusses similar philosophical resolutions to the paradoxes and presents similar decision schemes. Yudkowsky (2010) shows TDT succeeding in the original Newcomb's problem. Unfortunately, deciding exactly when and where to put logical nodes, and what conditional probabilities to place on them, is not yet an algorithmic process.

How would TDT look if instantiated in a more mature application? Given a very large and complex network, TDT would modify it in the following way: It would investigate each node, noting the ones that were results of instantiated calculations. Then it would collect these nodes into groups where every node in a group was the result of the same calculation. (Recall that we don't know what the result is, just that it comes from the same calculation.) For each of these groups, TDT would then add a logical node representing the result of the abstract calculation, and connect it as a parent to each node in the group. Priors over possible states of the logical nodes would have to come from some other reasoning process, presumably the one that produces causal networks in the first place. Critically, one of these logical nodes would be the result of TDT's own decision process in this situation. TDT would denote that as the decision node and use the resulting network to calculate the best action by equation 1.1.

Some TDT-like results of mutual cooperation have been shown in programmatic frameworks. In this context, decision theories are actual computer programs whose input is their opponent's source code, and whose output is "Cooperate" or "Defect." Tennenholtz (2004) introduced this idea and demonstrated that the formal version of "Cooperate if and only if my opponent has the same source code" cooperates with itself. This game-theoretic work is continued in Monderer and Tennenholtz (2006) and Fortnow (2009). Furthermore, it has been demonstrated in Slepnev (2010) that the strategy "Cooperate if and only if I can formally prove that my opponent cooperates" will cooperate with itself, given enough time to search through proofs. This is true even if the source code is not identical, and relies on the bounded version of Löb's theorem. Yet this is not a true instantiation of TDT, as it "wastes" cooperation against agents who *always* cooperate. LaVictoire (forthcoming) makes some progress in finding simple algorithms that more closely match our expectations of TDT, but whether a formalization of TDT exists in this context is still an open question.

## 7. Open Problems

TDT is a step towards a better decision algorithm. Solving Newcomblike problems is a good first step, but many more well-defined problems and ill-defined problems remain to be solved and clarified respectively. Some specific examples are:

**Logical uncertainty.** In the calculator problem, it is given that the prior probability of the third digit of $758439 \times 990452$ being odd is .5. This is obviously true. But how do we form these priors in general? All mathematical and logical theorems are necessarily true results of axioms. But due to bounded computational resources humans and other reasoning algorithms don't automatically know which results hold. What probabilities should we assign to conjectures, in general? Recent works on this topic include Gaifman (2004) and Haenni (2005).

**Inferring cause.** Like CDT, TDT must be given a causal network in order to algorithmically choose the correct decision. How can we derive causal graphs (which include ourselves as deciders!) from generic data? Some work on deriving causal structure from data is presented in Pearl (2009, chap. 2).

**Placing logical nodes.** As a companion problem to inferring cause, algorithmic methods are needed to infer when nodes represent the same or similar types of calculations, and further what conditional probabilities to assign. Giving such meaning to nodes in Bayesian networks is reminiscent of the symbol grounding problem in AI; rules for processing nodes are clear, while the meaning of the nodes themselves is less so.

**Refining probabilistic graphs.** It is possible that the universe is deterministic; therefore all probability assignments could be considered uncertainty in the logical implications of the laws of physics. Where should we consider a cause to be probabilistic, and where should we consider it logical uncertainty?

A Comparison of Decision Algorithms on Newcomblike Problems